\documentclass[review,1p]{elsarticle}
\usepackage{lineno,hyperref}
\usepackage{graphicx}
\usepackage{amsmath}
\usepackage{amssymb}
\usepackage{bm}
\usepackage{booktabs}
\usepackage{multirow}
\usepackage{colortbl}
\usepackage{algorithm}
\usepackage{algorithmic}
\usepackage{caption}
\usepackage{subcaption}
\usepackage[numbers]{natbib}
\modulolinenumbers[5]
\newtheorem{prop}{Proposition}
\newtheorem{definition}{Definition}[section]

\journal{Journal of Knowledge-Based Systems}

\begin{document}

\begin{frontmatter}

\title{Neighbor Enhanced Graph Convolutional Networks for Node Classification and Recommendation}

%% Group authors per affiliation:
\author[a1]{Hao Chen}
\ead{chh@buaa.edu.cn}
\author[a6]{Zhong Huang}
\ead{mad_hz@163.com}
\author[a4]{Yue Xu}
\ead{guyue.yuexu@alibaba-inc.com}
\author[a5]{Zengde Deng}
\ead{dengzengde@gmail.com}
\author[a2]{Feiran Huang}
\ead{huangfr@jnu.edu.cn}
\author[a3]{Peng He}
\ead{paulhe@tencent.com}
\author[a1]{Zhoujun Li\corref{mycorrespondingauthor}}
\cortext[mycorrespondingauthor]{Corresponding author}
\ead{lizj@buaa.edu.cn}

\address[a1]{State Key Laboratory of Software Development Environment, Beihang University, Beijing 100191, China.}
\address[a3]{WeChat, Tencent Inc., Shenzhen 518000, China}
\address[a6]{School of Information and Mathematics, Yangtze University, Jingzhou 434023, China}
\address[a4]{Alibaba Group, Hangzhou 310052, China}
\address[a5]{Cainiao Network, Hangzhou 311101, China}
\address[a2]{College of Cyber Security, Jinan University, Guangzhou 510632, China.}

\begin{abstract}
The recently proposed Graph Convolutional Networks (GCNs) have achieved significantly superior performance on various graph-related tasks, such as node classification and recommendation. However, currently researches on GCN models usually recursively aggregate the information from all the neighbors or randomly sampled neighbor subsets, without explicitly identifying whether the aggregated neighbors provide useful information during the graph convolution. In this paper, we theoretically analyze the affection of the neighbor quality over GCN models' performance and propose the Neighbor Enhanced Graph Convolutional Network (NEGCN) framework to boost the performance of existing GCN models. Our contribution is three-fold. First, we at the first time propose the concept of neighbor quality for both node classification and recommendation tasks in a general theoretical framework. Specifically, for node classification, we propose three propositions to theoretically analyze how the neighbor quality affects the node classification performance of GCN models. Second, based on the three proposed propositions, we introduce the graph refinement process including specially designed neighbor evaluation methods to increase the neighbor quality so as to boost both the node classification and recommendation tasks. Third, we conduct extensive node classification and recommendation experiments on several benchmark datasets. The experimental results verify that our proposed NEGCN framework can significantly enhance the performance for various typical GCN models on both node classification and recommendation tasks.
\end{abstract}


\end{frontmatter}

%\linenumbers

	\section{Introduction}
	Unlike traditional data mining systems, where the main purpose is to predict with the characters of single instances, modern systems record the connections between instances such as citations between papers or interactions between users and items to formulate these records as graph-structure data. Consequently, the Graph Convolutional Networks (GCNs), which can simultaneously tackle the graph topology and the node feature, have soon been popular and achieved state-of-the-art performances on various graph-based tasks such as node classifications~\cite{kipf2016gcn,velivckovic2017gat,zhang2018gaan,huang2018asgcn,chen2018fastgcn,zeng2020graphsaint,abu2019mixhop} and recommendation~\cite{fan2019meirec,zhao2019intentgc,he2020lightgcn,wang2019ngcf}. 
	
    Generally, the core of GCN models is to recursively update the representation of each node by aggregating the message passed from neighbors. To this end, GCN~\cite{kipf2016gcn} first proposes the concept of spectral graph convolution, which utilizes the whole graph connections~(\emph{i.e.}, adjacent matrix) to aggregate neighbors' information. Following its up, GAT models~\cite{velivckovic2017gat,zhang2018gaan} adapt the attention mechanism to dynamically adjust the aggregation weights of each neighbor. MixHop and JK-Nets~\cite{abu2019mixhop,xu2018jknet} introduce mixed graph convolution layers to simultaneously collect the information of the neighbors from different hops away. Considering the large computational cost of the recursive graph convolution, SGC~\cite{wu2019sgc}, NRGCN~\cite{chen2021nrgcn}, LightGCN~\cite{he2020lightgcn} and Gunets~\cite{gunets} omit the transformation function and the non-linear activation function to accelerate the graph convolutional process.
	
	Meanwhile, another line of works implements the graph convolution in a spatial manner, importing the sampling and clustering strategy to facilitate the graph convolutions on large graphs. GraphSAGE~\cite{hamilton2017graphsage} at the first time introduces the node-wise sampling and aggregation pipeline of the spatial graph convolution. For each node, GraphSAGE recursively samples and aggregates the 1-hop neighbors, 2-hop neighbors, and so on. Since the recursive sampling will inevitably involve exponentially increasing neighbors, FastGCN~\cite{chen2018fastgcn} proposes layer-wise sampling to randomly drop the nodes of the whole graph, avoiding the exponential growth. Besides, ASGCN combines node-wise and layer-wise sampling. Cluster-GCN~\cite{chiang2019cluster} and GraphSAINT~\cite{zeng2020graphsaint} split the whole graph into several subgraphs to tackle extremely large-scale graphs.
	
	However, although existing GCN models have set new standards on many benchmark tasks, they suffer from the following two main pitfalls.
	
	\smallskip\noindent\textbf{Ambiguous Neighbor Quality Evaluation.}
	As studied in Graph Adversarial Attack researches~\cite{dai2018adversarial,zugner2018adversarial,zugner2019adversarial}, the quality of the neighbors will impact the performance of GCN models intensively. Nonetheless, current models do not utilize explicit methods to evaluate the quality of the neighbor. For example, GAT models~\cite{velivckovic2017gat,wang2019kgat,wang2019hgat,zhang2018gaan} use the attention mechanism to assign different weights to different neighbors. However, attention mechanisms are proposed to focus on the informative nodes, which cannot exactly identify the quality of neighbors. Besides, the attention mechanism will import a large number of additional operations, however, the attention weights cannot be used to enhance the performance of other non-attention GCN models. 
    Noticeably, the recent studies on graph adversarial attacks indicate that GCN models~(including GATs) are vulnerable to adversarial attacks which debase the quality of neighbors by adding disturbing neighbors or deleting informative neighbors. \emph{This motivates us to rethink how to explicitly evaluate the neighbor quality, making most GCN models~(including GATs) can benefit from the neighbor quality evaluation and enhancement}.
	
	\smallskip\noindent\textbf{Potential Disturbing Neighbors.}
	For node classification tasks, GCN models rely on the homophily hypothesis~\cite{mcpherson2001homophily}, meaning that the connected nodes usually have similar features or characters. However, in real datasets, there exist some nodes that do not have enough similar neighbors while disturbing by much more dissimilar neighbors. The potential disturbing neighbors will influence the performance of GCN models on node classification tasks. 
	Moreover, in the recommendation scenario, the potential disturbing neighbors are more difficult to recognize, since the interactions between users and items are implicit. Existing models usually employ intuitive methods to evaluate the similarity between users and items, \emph{i.e.}, comparing the weights of the interactions. However, such intuitive evaluation methods are susceptible to popular items while cannot provide theoretical guarantees.
	\emph{Considering both node classification and recommendation tasks, it is necessary to develop a general theoretical framework to tackle the potential disturbing neighbors for these two different but correlated tasks.}

	\smallskip\noindent\textbf{Our Work.}
	In this paper, we consider the tasks of node classification and recommendation. We propose the general framework Neighbor Enhanced Graph Convolutional Networks(NEGCN) to refine the graph structure before the training of GCN models to improve their learning performance. The main contributions are summarized as follows.
	\begin{itemize}
		\item This paper for the first time proposes a general theoretical framework for both node classification and recommendation tasks. Specifically, for node classification tasks, we propose three propositions to theoretically analyze how the neighbor quality affects the performance of GCN models. 
		
		\item Based on the three proposed propositions, we introduce the neighbor evaluation measurement and present specially designed neighbor evaluation methods and graph refinement policies to increase the neighbor quality so as to enhance the GCN models' performance on the node classification and recommendation tasks. Specifically, we propose an efficient edge classifier to predict whether a neighbor is a useful neighbor. Then we modify the graph structure by filtering the useless 1-hop neighbors and useful 2-hop neighbors as extra useful 1-hop neighbors.
		
		\item Extensive experiments results for both node classification and recommendation tasks verify that our proposed NEGCN framework is generally powerful in boosting various popular GCN models for node classification tasks, including spectral GCN, SGC, and GAT and sampling-based GraphSAGE and ASGCN. Additionally, experiments on recommendation datasets further verify the universality of NEGCN.
	\end{itemize}

    This article extends the short proceeding paper~\cite{lagcn} in theoretical support and applications. First, we employ mutual information to generally model and explain the neighbor quality evaluation.  Second, equipped with the theoretical guarantees, we generalize the neighbor quality evaluation and the graph refinement onto the recommendation scenario. Finally, extensive experiments and comparison that adapts NEGCN on recommendation tasks are proposed to present the universality of our model.
    
    The whole paper is organized as follows. In Section 2, we review the recent researches in node classification and recommendation tasks. Section 3 and Section 4 propose the neighbor quality evaluation concept and describe how to enhance GCN models for node classification and recommendation. In Section 5, we present the experimental results on the five benchmark datasets and analyze the results. Finally, we conclude our paper in section 6.
    
	\section{Related Work}
	\subsection{Node classification.}
	Existing GCN models for node classification can be divided into two main categories~\cite{gilmer2017nmp,zhou2018survey,zhang2018survey,wu2019survey}: 1) spectral~(transductive) GCN models and 2) spatial~(inductive) GCN models.
	Spectral GCN models define graph convolutions based on the spectral representation of a graph. The first spectral GCN model is proposed in~\cite{bruna2013spectral}, which defines the graph convolution operation in the Fourier domain. Later, the localized filters and Chebyshev expansion~\cite{duvenaud2015convolutional,niepert2016learning,defferrard2016convolutional} are proposed to avoid the need for computing the eigenvectors of the Laplacian matrix. Next, \cite{kipf2016gcn}~simplifies the convolution operation via a localized first-order approximation. Recently, to accelerate the convolution process, SGC~\cite{wu2019sgc} removes the nonlinear transition function between consecutive GCN layers while still reaching a comparable performance to the traditional GCN. JK-Nets and MixHop\cite{xu2018jknet,abu2019mixhop,wu2021enhancing} define multiple-hop graph convolution to directly access the neighbors multi-hops away. However, most of the learned filters in spectral GCN models depend on the whole graph structure, which is transductive and computationally inefficient.
	
	The spatial (inductive) GCN models propose mini-batch training on graphs, which operates on spatially connected neighbors~\cite{zhou2018survey,hamilton2017graphsage}. In particular, GraphSAGE~\cite{hamilton2017graphsage} proposes the sampling and aggregation framework. ASGCN~\cite{huang2018asgcn} controls the sampling size of the training for GCNs through adaptive sampling. \cite{chen2017stochastic,chen2018fastgcn,zeng2020graphsaint,chiang2019cluster} further develop graph sampling, graph clustering and implement multiple sampling in an inductive GCN framework. 
	However, none of the GCN models mentioned above explicitly evaluates the quality of the neighboring nodes or investigates how to enhance the accuracy of GCN models by refining the graph structure.
	Another line of studies generalizes the graph convolution with the attention mechanism. Gat and Gaan\cite{velivckovic2017gat,zhang2018gaan,zhang2021graph} dynamically assign the aggregation weights with the multi-head attention mechanism. However, the attention mechanism can not gain valuable information from other non-neighboring nodes. Moreover, training the attention layers requires an extensive amount of extra computational overhead. Meanwhile, other GCN models with different architectures cannot take advantage of the trained attention weights.

	Besides the supervised methods, unsupervised learning methods can learn the graph representation without using the label information. Early methods for graph representations learning are mainly based on matrix-factorization approaches \cite{belkin2002laplacian,ahmed2013distributed,cao2015grarep,ou2016asymmetric}. With the popularity of the word embedding approach \cite{mikolov2013distributed}, many methods combine the random-walk and word embedding to generate the unsupervised embeddings \cite{perozzi2014deepwalk,grover2016node2vec}. Recently, \cite{velivckovic2018dgi} utilize graph convolution methods to generate the node embeddings.

	\subsection{Recommendation}
    The mainstream of graph-based recommendation is constructing the user-item interaction graphs and then trying to learn the implicit relations between users and items.
    Traditionally, matrix factorization methods~\cite{koren2009MF,rendle2012bprmf,rao2015grmf} propose the groundwork of embedding-based recommendation which learns the embedding for each user and each item by reconstructing the user-item interaction graphs. Motivated by this, network embedding methods~\cite{perozzi2014deepwalk,grover2016node2vec,zhao2017meta,dong2017metapath2vec,fu2017hin2vec} perform several random-walk strategies to extract the relations between users and items and then utilize the skip-gram models~\cite{mikolov2013word2vec} to turn them into embeddings. 
    
    Since GCN models have presented impressive performances on various graph-related tasks, there is a growing number of researchers studying how to generalize GCN models~\cite{zhao2019intentgc,fan2019meirec, hu2018leveraging, fan2019graph, yin2019deeper, salamat2021heterographrec} to tackle user-item recommendation problems.
	In particular, IntentGC~\cite{zhao2019intentgc} accelerates neighborhood feature propagation by introducing the vector-wise graph convolution to avert unnecessary interactions between features.
	KGAT~\cite{wang2019kgat} expands the graph attention networks to compute the hidden states of heterogeneous graphs. 
	Dual Graph Attention Networks~\cite{wu2019dual} instead utilize the multi-arm bandit to explore the user interests with graph attention. MEIRec~\cite{fan2019meirec} leverages the long-short-term-memory~(LSTM) Networks to capture the sequential correlation between different types of correlations. 

	Recent recommendation models also investigate several policies to evaluate the node similarities to achieve better recommendation performance.
	Among these policies, one widely used policy is the first-order interaction weights. In particular, many models consider a user to be more correlated to an item if the user clicks or purchases the item more frequently, related works including Dual Graph Attention Networks~\cite{wu2019dual}, MEIRec~\cite{fan2019meirec}, KGCNs~\cite{Wang2019graph,wang2019knowledge}, etc.
	Another orientation of methods prefers the second-order interaction weights, which measures the similarity of two nodes by comparing the occurrence frequency of neighborhood structure~\cite{goyal2018graph}. IntentGC~\cite{zhao2019intentgc} infers neighbors' similarity by counting the commonly visited neighbors.

	However, these works do not investigate explicit evaluation metrics to measure the importance of a given neighbor but only provide empirical importance evaluation.
	
	\section{Preliminaries}
	\subsection{Notations}
	\textbf{Node Classification.} We consider the node classification problem on a directed graph $\textbf{G}=(\textbf{V},\textbf{E})$ with $N$ nodes and $E$ edges. 
	The edge between node $ u,v \in \textbf{V}$ is denoted as $(u,v)\in\textbf{E}$. The binary adjacency matrix with self-loop edges is denoted as $\textbf{A}\in\mathbb{R}^{N\times N}$.
	Let $\textbf{X} = \{\textbf{x}_1, \textbf{x}_2, \cdots, \textbf{x}_N\}$ denotes the feature of the nodes and $Y=\{y_1, y_2, \cdots, y_N\}$ denotes the labels of all the nodes. Let $\mathcal{N}_v$ denote the set of all $ 1 $-hop neighbors of node $v$ and $n_v=|\mathcal{N}_v|$ denote the number of $ 1 $-hop neighbors of node $v$. 
	
	\textbf{Recommendation.} We suppose the sets of users and items are $\mathbf{U}$ and $\mathbf{I}$, respectively. $u\in\mathbf{U}$ denotes the user index, $i\in\mathbf{I}$ denotes the item index. Index $j$ denotes the index that is not distinguished between user and item. The matrix $\mathbf{R}\in\mathbb{R}^{N\times M}$ denotes the observed user-item interactions, where $N$ and $M$ is the quantity of the users and the items, respectively. Specifically, $\mathbf{R}_{ui}=1$ if the interaction between user $u$ and item $i$ is observed, while $\mathbf{R}_{ui}=0$ denotes the unobserved interactions. 
    Since in recommendation situations, users and items do not have node features as in node classification scenarios. Thus, multiple collaborative filtering~(CF) embedding methods are utilized to train their embeddings. Especially, $\textbf{x}_u$ denotes the CF embedding of user node $u$, and $\textbf{x}_i$ denotes the CF embedding of item node $i$. Besides, we mathematically define the adjacent matrix under the recommendation situation as  
    \begin{align}
    \textbf{A} = 
    \left[
        \begin{matrix}
           \textbf{0} & \textbf{R}  \\
           \textbf{R}^\top & \textbf{0} 
          \end{matrix}
    \right] \in \mathbb{R}^{(N+M)\times (N+M)}.
    \end{align}
	
	\subsection{Graph Convolutional Networks}
	In this subsection, we review the two most popular branches of the graph convolutional networks: the spectral graph convolutional networks and the spatial graph convolutional networks. 

    Generally speaking, the spectral branch of methods~\cite{kipf2016gcn,chen2017stochastic,abu2019mixhop,velivckovic2017gat} utilizes the adjacent multiplier to compute the graph convolution, while the spatial branch of methods~\cite{hamilton2017graphsage,chen2018fastgcn,huang2018asgcn} propose the sampling and aggregation framework to pass the message from neighbors to the central node step by step. The Graph Convolutional Network~(GCN) proposed in~\cite{kipf2016gcn} is the first spectral graph model to adjust the convolutional layers in CNN into graph structure data. The core structure of the GCN model is the first-order graph convolution layer that is defined as,
	\begin{equation}
	\textbf{H}^{(l+1)} = \sigma(\hat{\textbf{A}}\textbf{H}^{(l)}\textbf{W}^{(l)}),
	\end{equation}
	where the convolutional multiplier $\hat{\textbf{A}}=\textbf{D}^{-\frac{1}{2}}\textbf{A}\textbf{D}^{-\frac{1}{2}}$ denotes the normalized adjacency matrix with $\textbf{D}$ as the diagonal degree matrix. $\textbf{H}^{(l)}$ denotes the node embeddings after $l$-times convolution, $\textbf{W}^{(l)}$ denotes the transforming mapping matrix of the $l$-th graph convolutional layer, and $\sigma(\cdot)$ denotes the non-linearity activation function. The initial node embedding at the first layer $\textbf{H}^{(0)}$ is initialized as the raw feature $\textbf{X}$. After K steps of the first-order graph convolution, we have $\textbf{H}^{(K)}$ as the final embedding of a $K$-layer GCN model. 
	
	Since original GCN models have to dynamically compute the intermediate embeddings before computing the final embeddings $\textbf{H}^{(K)}$, whereas causing severe computational inefficiency for the repetitive convolution process. To mitigate the computational burden, the Simplified Graph Convolutional Networks~(SGCs)~\cite{wu2019sgc} simplify the complicated graph convolutional layers by omitting the non-linearity including the non-linear activation function $\sigma(\cdot)$ as well as the transforming mapping matrix $\textbf{W}^{(l)}$. Interestingly, SGC theoretically and experimentally proves that the skipping of the non-linearity does not affect the predicting performance of the graph convolutional networks. As such, the final embedding matrix given by a $K$-layer GCN can then be simplified as,
	\begin{equation}
	\textbf{H}^{(K)} = \hat{\textbf{A}}^K\textbf{X}.
	\label{equ:sgc}
	\end{equation}
	
	GraphSAGE~\cite{hamilton2017graphsage} is the first spatial graph convolutional models, which proposes the sampling and aggregation framework. Since spectral convolution has to get access to the whole adjacent matrix to obtain the graph convolution these models cannot deal with the inductive learning problem which requires the trained models to generalize to unseen nodes with the newly observed subgraphs. To this end, GraphSAGE redefine the graph convolution in the spatial domain to extract neighbors' message with the following formula,
	\begin{equation}
	\textbf{h}_{v}^{(l+1)} = \sigma(\text{MAP}_{l}(\text{AGGREGATE}_{(l)} (\{\textbf{h}^{(l)}_u \text{, for }u\in\mathcal{N}_v\})),
	\label{equ:sage}
	\end{equation}
	where ``MAP'' refers to the linear transformation that works the same as the transforming mapping matrix $\textbf{W}^{(l)}$. ``AGGREGATE'' denotes various neighbor aggregation methods such as mean pooling, max pooling, summation, and other LSTM neural networks. Empirically, GraphSAGE opts for mean pooling as the default aggregation function, then the embedding of node $v$ at the $l+1$ layer of these models can be computed as,
	\begin{equation}
	\textbf{h}_{v}^{(l+1)} = \sigma(\textbf{FC}_{\theta_{(l)}}(\sum_{u\in{\mathcal{N}_v^*}} w^{(l)}_{u,v} \textbf{h}^{(l)}_u )),
	\label{equ:agg}
	\end{equation}
	where $\textbf{FC}_{\theta_{(l)}}$ represents the fully-connect layer parameterized by $\theta_{(l)}$; $w^{(l)}_{v,u}$ denotes the aggregation weight of the neighboring nodes $u\in{\mathcal{N}_v}$. Especially, for GraphSAGE-mean-pooling, $\mathcal{N}_v^*$ denotes the sampled subset of the original neighbors set $\mathcal{N}_v$, and the aggregation weights can be given by the mean pooling structure, where $w_{u,v}^{(l)} = 1/|\mathcal{N}_v|$.
	
	Another line of works focus on applying the attention mechanism in NLP processing~\cite{vaswani2017attention} to dynamically assign the above aggregation weights. Typically, the computation contains two steps: computing attention coefficients and normalizing the attention weight. Graph Attention Networks~(GAT)~\cite{velivckovic2017gat} perform self-attention between the central node and its 1-hop neighbors as,
	\begin{equation}
	    w^{(l)}_{u,v} = \frac{\exp(\sigma(\textbf{a}^\top \cdot [\textbf{FC}_{\theta_{(l)}}(\textbf{h}_u^{(l)})\parallel \textbf{FC}_{\theta_{(l)}}(\textbf{h}_v^{(l)})]}{\sum_{k\in\mathcal{N}_v}\exp(\sigma(\textbf{a}^\top \cdot [\textbf{FC}_{\theta_{(l)}}(\textbf{h}_k^{(l)})\parallel \textbf{FC}_{\theta_{(l)}}(\textbf{h}_v^{(l)})]},
	\end{equation}
	where $\textbf{a}$ denotes the attention weight vector.
	
	\section{Neighbor Enhanced Graph Convolutional Networks}
	In this section, we first introduce the concept of neighbor quality. Then, we specify how to evaluate the neighbor quality on the node classification task by proposing the positive ratio. Furthermore, we discuss the construction of the edge classifier to evaluate the neighbor importance and analyze how the positive ratio impacts the node classification performance. Especially, we present how to refine the homogeneous graphs into neighbor enhanced graphs~(NE-graphs) for better node classification performance with the trained edge classifiers. Finally, we provide the approach to adapt the neighbor quality evaluation to enhance the recommendation tasks.
	
	\subsection{Neighbor Quality Evaluation}
    As described in~\cite{li2018laplacian}, the substantial enhancement of Graph Convolutional Networks is to describe the node with its neighbors. Intuitively, it is possible to define the quality of one neighbor by evaluating how well can we infer the node's characters from this neighbor. To stabilize the evaluation of the neighbor quality, we opt for mutual information to quantitatively study how much information can a neighbor provide to infer the central node.
    
    Let $v$ denote the central node. We randomly choose a neighbor $u$ from the neighbor set of $v$. Let the random variable $\textbf{u}$ denote the feature of an arbitrary node in $\mathcal{N}_u$, then the distribution of the random variable $\textbf{u}$ is $P_\textbf{u} = P(\textbf{u} = \textbf{x}_u)$, where $\textbf{x}_u$ is the outcome feature of node $u$. Analogously, we assume the random variable $\textbf{v}$ to describe the central node $v$, then the distribution of $\textbf{v}$ is $P_\textbf{v} = P(\textbf{v} = \textbf{x}_v)$,  where $\textbf{x}_v$ is the outcome feature of node $v$. After defining these notations, their mutual information $I(\textbf{u},\textbf{v})$ is the KL-divergence between the joint distribution $P_{\textbf{u},\textbf{v}} = P(\textbf{u} = \textbf{x}_u, \textbf{v} = \textbf{x}_v)$ and the product of marginal distributions $P_\textbf{u} \otimes P_\textbf{v}$:
    \begin{align}
        \begin{split}
            &I(\textbf{u},\textbf{v}) = D_\text{KL}(P_{\textbf{u},\textbf{v}}\parallel P_\textbf{u} \otimes P_\textbf{v})\\
            \overset{(a)}{\geq} &\sup\limits_{T\in \mathcal{T}} \left\{ \mathbb{E}_{\textbf{x}_u,\textbf{x}_v\thicksim P_{\textbf{u},\textbf{v}}}[T(\textbf{x}_u,\textbf{x}_v)] - \mathbb{E}_{\textbf{x}_u\thicksim P_{\textbf{u}}, \textbf{x}_v'\thicksim P_{\textbf{v}}}[e^{T(\textbf{x}_u, \textbf{x}_{v'})-1}]
            \right\},
        \end{split}
        \label{eq:kl}
    \end{align}
    where $(a)$ follows from $f$-divergence representation based on KL-divergence~\cite{mohamed2018mutual}; random variable $\textbf{v}'$ denotes the feature associate with an arbitrary node; $T\in \mathcal{T}$ is an arbitrary function that maps a pair of features to a real value, which reflecting the correlation of two features. After replacing the $f$-divergence in Eq.~(\ref{eq:kl}) with a GAN-like divergence, Eq.~(\ref{eq:kl}) is written as follows,
    \begin{align}
        \begin{split}
            I_{\text{GAN}}(\textbf{u},\textbf{v}) \geq &\sup\limits_{T\in \mathcal{T}}\{
            \mathbb{E}_{P_{\textbf{u},\textbf{v}}}[\text{log}\sigma(T(\textbf{x}_u,\textbf{x}_v))] \\&+ \mathbb{E}_{P_\textbf{u},P_{\textbf{v}}}[\text{log}(1-\sigma(T(\textbf{x}_u,\textbf{x}_{v'})))]
            \},
        \end{split}
        \label{eq:gan}
    \end{align}
    where $\sigma(\cdot)$ is the sigmoid function. 
    Specifically, in $I_{GAN}(\textbf{u},\textbf{v})$, the first term reflects the correlation between the neighbor $u$ and the central node $v$, meaning to what extend the neighbor $u$ describe the central node $v$. The second term evaluates the difference between the sampling node and the other node, which estimates the particularity of the neighbor $u$. In practice, we cannot go over the entire functional space to evaluate the exact value of $I_{\text{GAN}}$. As a trading-off, we only preserve the first item to reflect the importance of a given neighbor. Thus the core of identifying the importance of a neighbor comes to be the training of the mapping function $T(\cdot)$. In the following subsections, we respectively describe that how to train powerful $T(\cdot)$ for node classification and recommendation tasks, respectively.
	
	\begin{figure*}[t]
		\begin{center}
			\centerline{\includegraphics[width=\linewidth]{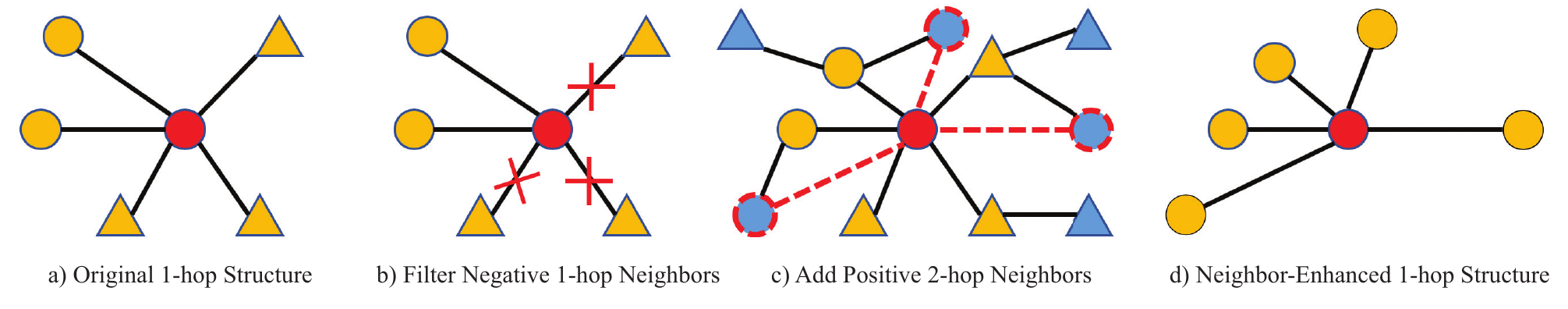}}
			\vskip -1em
			\caption{Illustration of the neighbor-enhancing process for the node classification task. The red, yellow, and blue nodes denote the central nodes, the 1-hop neighbors, and the 2-hop neighbors, respectively. The circles denote the same-label~(positive) neighbors, while the triangles denote the different-label~(negative) neighbors. (a) The original 1-hop neighbors of the central node. (b) The filtering process, i.e., deleting the 1-hop neighbors from the central node, which have different labels. (c) The adding process, i.e., connecting the selected 2-hop neighbors to the central node, which has the same label. (d) The refined 1-hop neighbors of the central node in the node classification task~(neighbor enhanced graph).
			}
			\label{fig:framework}
		\end{center}
		\vskip -2 em
	\end{figure*}
	\subsection{Positive Ratio for Node Classification}
	Given a central node, we can classify its neighbors into the positive neighbors and the negative neighbors, which are defined as follows.
	\begin{definition}\label{def1} 
		Given the central node $v$, the positive neighbor set $\mathcal{N}^+_v$ is defined as the subset of neighbors that have the same label as $y_v$. The negative neighbor set $\mathcal{N}^-_v$ is defined as the subset of neighbors that have different labels from $y_v$. 
	\end{definition}
	Based on the definition, we rewrite the update function of node $v$ given in~\eqref{equ:agg} as,
	\begin{equation}
	\textbf{h}_{v}^{(l+1)} = \sigma (\textbf{FC}_{\theta_{(l)}}(\sum_{u\in{\mathcal{N}^+_v}} w^{(l)}_{v,u} \textbf{h}_{u}^{(l)} + \sum_{u\in{\mathcal{N}^-_v}} w^{(l)}_{v,u} \textbf{h}_{u}^{(l)})).
	\end{equation}
	According to~\cite{wu2019sgc}, the non-linearity between consecutive GCN layers are not necessary for performing the graph convolution. Instead, by removing the non-linearity, SGC can achieve comparable performance as GCN and 10 times faster than GCN. 
	Thus, we drop the non-linearity and rewrite the updating function of node as,
	\begin{equation}
	\textbf{h}_v^{(l)} = \frac{1}{n_v}(\sum_{u\in{\mathcal{N}^+_v}}  \textbf{h}_{u}^{(l-1)} + \sum_{u\in{\mathcal{N}^-_v}} \textbf{h}_{u}^{(l-1)}).
	\label{equ:sgc_agg}
	\end{equation}
	Helped by the linear updating function, the embeddings after different times of convolutions locates at the same hidden space. Analogous to the one-vs-all classification in multi-class prediction, we turn the multi-class node classification problem to the binary classification, where classifying the central node to its correct label as one class and classifying the central node to wrong labels as the other class. Formally, we make following assumptions:
	\begin{itemize}
		\item Assuming the linear mapping $F(\cdot)$ that maps the hidden embedding after $(l-1)$-times graph convolution to a real number, i.e., $F(\textbf{h}_u^{(l-1)})$, we classify the node $u$ into the positive class when $F(\textbf{h}_u^{(l-1)}) > \tau$, or negative class otherwise, where $\tau$ is the threshold of the one-vs-all classification.
		\item We assume the mapping of $F(\cdot)$ to obey a Gaussian mixture distribution, where  $F(\textbf{h}_u^{(l-1)}){ iid \atop \sim} Norm(\mu^+, \sigma^2)$ for $u \in \mathcal{N}^+_i$ and $F(\textbf{h}_u^{(l-1)}){ iid \atop \sim} Norm(\mu^-, \sigma^2)$ for $u \in \mathcal{N}^-_i$. 
		Reasonably, we assume that $\mu^+$ is larger than $\tau$, and $\mu^-$ is smaller than $\tau$.
	\end{itemize}
	Following above assumptions, $F(\textbf{h}_v^{(l)})$ can be written as follows,
	\begin{equation}
	F(\textbf{h}_v^{(l)}) = \frac{1}{n_v}
	(\sum_{u\in{\mathcal{N}_v^+}} F(\textbf{h}_u^{(l-1)})
	+ \sum_{u\in{\mathcal{N}_v^-}} F(\textbf{h}_u^{(l-1)}).
	\end{equation}
	We compute the expectation of the mapping results $F(\textbf{h}_v^{(l)})$ as follows,
	\begin{equation}
	E_{origin} = \dfrac{1}{n_v}(n^+_v \mu^+ + n^-_v \mu^-) = r_v \mu^+ + (1-r_v)\mu^-,
	\end{equation}
	where $r_v = n_v^+/(n_v^+ + n_v^-) = n_v^+/n_v$ denotes the positive ratio of node $v$. 
	Observing from the formula of expectation, we can enlarge the expectation of the correctly classifying possibility by increasing the positive ratio~$r_v$ for node $v$. This observation leads to the following proposition.
	\begin{prop}\label{the:origin}
		The probability of classifying the nodes $v \in \textbf{V}$ into the correct class will increase if we can enlarge the positive ratio of the neighbors of node $v$. Thus the total node classification performance will increase when we enhance the whole graph to have larger global positive ratio $R=\frac{1}{N}\sum_{v\in\textbf{V}} \ n_v^+/(\sum_{v\in\textbf{V}}n_v)$.
	\end{prop}
	
	\subsection{Edge Classifier}
	According to Definition~\ref{def1}, in order to identify whether node $u$ is a positive neighbor of node $v$, when the labels of $u$ and $v$ are unknown, such as in the testing set. 
	The most straightforward way is to first predict the labels of node $u$ and node $v$, and then identify whether they are the same. 
	However, this leads to a classification problem with $C\times C$ classes, which is complicated and expensive to solve, especially when $ C $ is large. 
	To this end, we now build an edge classifier to convert the above classification problem into a binary classification problem. 
	Specifically, we define the edges between the same-label nodes as \textit{positive edges} and the edges between the node with different labels as \textit{negative edges}. 
	
	Given a pair of nodes $(u, v)$, the edge classifier $\mathcal{E}$ is trained to return a continuous value between 0 and 1 named $\hat{y}_{u,v}$ that reflects the possibility that the labels of $u$ and $v$ are the same.
	The edge classifier is trained through the adjacent matrix $\textbf{A}$, the feature of the nodes $\textbf{X}$, and the label-information of the nodes in the train set $Y_{train}$.
	The edge classifier should be easily computed and satisfy the commutative property which means exchanging the order of the edge will not influence the prediction (i.e., $\mathcal{E}(u,v) = \mathcal{E}(v,u)$).
	
	In this paper, we adopt multi-layer perception (MLP) layers to build the edge predictor. Given the input features of the two nodes as $\textbf{e}_u$ and $\textbf{e}_v$, we pass the concatenation of the absolute difference $|\textbf{e}_u - \textbf{e}_v|$, the summation $\textbf{e}_u + \textbf{e}_v$, and the Hadamard product $\textbf{e}_u \circ \textbf{e}_v$ of the input features into the edge classifier.
	Moreover, in order to accelerate the prediction process under high-dimensional features, we use a weighted matrix $\textbf{W}_e$ to project the input features into a space with lower dimensions~(i.e., $\hat{\textbf{e}}_u =\textbf{e}_u \textbf{W}_e$). 
	As such, the edge classifier can be written as
	\begin{equation}
    \mathcal{E}(u,v) = \textbf{MLP}\left(|\hat{\textbf{e}}_u - \hat{\textbf{e}}_v| \ \| \ (\hat{\textbf{e}}_u + \hat{\textbf{e}}_v) \ \| \ (\hat{\textbf{e}}_u \circ \hat{\textbf{e}}_v)\right),
	\end{equation}
	where $\|$ denotes concatenating two vectors and $\textbf{MLP}$ denotes the multi-layer perception with non-linear activation. 
	Using MLP layers in building the edge classifier is not the only possible solution, here we opt for MLP for easy-computing and satisfying the commutative property.
	
	\subsubsection{Graph Refinement for Node Classification Tasks.}
	The graph refinement for node classification tasks aims at using the edge classifier to enhance the origin graph structure to achieve a better node classification performance. Particularly, the modification process contains two parts: 
	1) \textbf{the filtering process}: as presented in Figure~\ref{fig:framework}(b), for every node in the dataset set, we use the trained edge classifier to predict whether its 1-hop neighbors are positive neighbors, then we delete all of its negative neighbors;
	2) \textbf{the adding process}: as shown in~Figure \ref{fig:framework}(c), for each node in the dataset set, we use the trained edge classifier to predict whether its 2-hop neighbors are positive neighbors, then we add edges between this node and its positive 2-hop neighbors, i.e., turning them into positive 1-hop neighbors, until the total number of 1-hop neighbors reaches a preset maximum neighbor number $n_{max}$.
	%\footnote{Practically, one can sort the positive neighbors according to the confidence level of the prediction results.} 
	Generally, if the edge classifier is good enough, such graph refinement can increase the graph's positive ratio $R$.
	
	\subsection{Theoretical Analysis}
	In this subsection, we describe the minimum requirements for the edge classifier to achieve a better performance by refining the original graph to NE-graph.
	
	For the \textbf{Filtering Process}, we delete all the predicted negative edges ($\hat{y}_{u,v} =0$) between the central node $v$ and its 1-hop neighbors. 
	After finishing the filtering process, we preserve the predicted positive neighbors containing the following two situations: 
	1) the predicted positive neighbors that are indeed positive, i.e., $y_{u,v} = 1$ and $ \hat{y}_{u,v} = 1$; 
	2) the predicted positive neighbors that are indeed negative, i.e., $y_{u,v} = 0$ and $\hat{y}_{u,v} = 1$. 
	We suppose the number of positive (or negative) neighbors for node $v$ before the filtering process $n_v^+$ (or $n_v^-$).
	Then, after the filtering process, the number of positive (or negative) neighbors for node $v$ turns to be $p\cdot n_v^+$ (or $q\cdot n_v^-$) where $p={\rm P}(\hat{y}_{u,v} =1|y_{u,v} = 1) $ and $q={\rm P}(\hat{y}_{u,v} =1|y_{u,v} =0)$.
	As such, the expectation of the map function $F(\textbf{h}_v^{(l)})$ is given as
	\begin{equation}
	E_{filter} = \frac{p\cdot n^+_v \mu^+ + q\cdot n^-_v \mu^-}{p\cdot n_v^+ + q\cdot n_v^-}.
	\end{equation}
	To make $E_{filter}$ larger than $E_{origin}$, the minimum requirement of the edge classifier should obey this proposition.
	\begin{prop}
		Given $p={\rm P}(\hat{y}_{u,v} =1|y_{u,v} = 1) $ and $q={\rm P}(\hat{y}_{u,v} =1|y_{u,v} =0)$, the neighbor-enhanced graph leads to a superior performance of the GCN models when the edge classifier satisfies $p > q$.
		\label{the:filter}
	\end{prop}

	For the \textbf{Adding Process}, for the nodes that do not have enough positive 1-hop neighbors, we employ the trained edge classifier to connect the central node to its predicted-positive 2-hop neighbors to increase the number of the positive 1-hop neighbors.
	Analogously, before the adding process, we assume the number of positive (or negative) neighbors to be $n_v^+$ (or $n_v^-$), $n_v'$ denotes the number of the neighbors we should add to node $v$. After enhancing node $v$ by adding $n_v'$ predicted positive neighbors, the positive-neighbor number of $v$ is $n_v^+ + p_{pre} \cdot n_v'$, and its negative-neighbor number is $n_v^- + (1-p_{pre}) \cdot n_v'$, where $p_{pre}={\rm P}(y_{u,v} = 1|\hat{y}_{u,v} =1)$ and $1-p_{pre}={\rm P}(y_{u,v} = 0|\hat{y}_{u,v} =1)$. 
	Thus, the expectation of the mapping function $F(\textbf{h}_v^{(l)})$ turns to be
	\begin{equation}
	E_{adder} = \frac{ (n^+_v+p_{pre}\cdot n_v') \mu^+ + (n^-_v + (1- p_{pre})\cdot n_v' )\mu^-}{n_v^+ + n_v^- + n_v'}.
	\end{equation}
	To make $E_{adder}$ larger than $E_{origin}$, the minimum requirement of the edge classifier should comply the following proposition.
	\begin{prop}
		Given $p_{pre}={\rm P}(y_{u,v} = 1|\hat{y}_{u,v} =1)$, the neighbor-enhanced graph leads to a superior performance of GCN models when the edge classifier satisfies $p_{pre} > r_v$. 
		\label{the:add}
	\end{prop}
	
	\subsubsection{Relationship with Existing GCN Models}
	The neighbor-enhanced graph has a higher positive ratio than the original graph, such that the disturbance from negative neighbors can be reduced. Existing GCN models can learn directly on the enhanced graph without changing their model architectures.
	On the other hand, the proposed NEGCN is promising to generate better performance than the attention mechanism in GAT~(which is verified in Sec.~\ref{sec:experiment}) due to the following reasons. 
	First, the attention mechanism can only filter the aggregated information from existing 1-hop neighbors, which may have little effect when the target node has very few positive 1-hop neighbors. Comparatively, NEGCN can connect positive 2-hop neighbors to the target node as added 1-hop neighbors to provide extra valuable information.
	Second, due to the lack of an explicit criterion, it is hard for the attention mechanism to accurately distinguish distracting neighbors and valuable neighbors, especially when the quality of the original graph is low. Comparatively, the NEGCN is more robust since it trains an edge classifier using the labels from the training set to explicitly split the positive neighbors and the negative neighbors.
	Third, the attention mechanism is trained together with the other convolution parameters using the full training set, which leads to an expensive computational cost. In particular, the computational complexity of training the attention layers is $\mathcal{O}(BHN^2)$, where $B$ is the number of the training batches, $H$ is the number of multi-head attention entities. 
	On the contrary, NEGCN requires a much lower computational cost. Besides, the graph refinement only needs to be done once, where the complexity of the filtering process is $\mathcal{O}(E)$ and the complexity of the adding process is $\mathcal{O}(N\cdot e^2)$, with $e$ being the average degree of the original graph.
	However, it is noteworthy that NEGCN can also boost the attention-based models~(GAT) to reach better performance, which is shown in Sec.~\ref{sec:experiment}.
	
	\subsection{Enhancing Recommendation Tasks}
	In node classification tasks, NEGCN can directly identify the positive and negative neighbors by observing the labels of the connected nodes and thus train the edge classifier to enhance the node classification graphs.
    By contrast, in the recommendation task, there is no explicit metrics to directly evaluate whether a neighbor is a positive or a negative one, meaning that we need to measure the similarity between two users (or items) based on their \textit{indirect~(implicit)} relationships. 
    Current researches commonly model the similarity between two nodes based on intuitive policies, e.g., considering the visiting times, or counting the shared viewed items~\cite{fan2019meirec, Wang2019graph, wu2019dual, wang2019knowledge}. However, these metrics will be affected by popular items or users. Besides, these metrics are unable to compare the similarity with the different user-item pairs. Therefore, it is worth exploring the definition of a metric to evaluate the neighbor quality for neighborhood aggregation in recommendation tasks.
	
    In $I_{GAN}(\textbf{u},\textbf{v})$, the first term reflects the correlation between the neighbor node $u$ and the central node $v$, meaning to what extend does the neighbor describe the central node $v$. The second term evaluates the difference between the sampling node and the whole user-item embeddings, which estimates the particularity of the node $u$. Since the feature of nodes are learned collaborative filtering embeddings, we parameterize $T(\textbf{a},\textbf{b})$ by computing their inner product as $T(\textbf{a},\textbf{b}) = \textbf{a}^\top \cdot \textbf{b}$. Besides, $\log(1-\sigma(\cdot))$ is a concave function, we can further simplify the second term of Eq.~(\ref{eq:gan}) with Jensen's inequality to obtain the following neighbor information evaluation function,
    \begin{equation}
        C(u,v) = \log\sigma(\textbf{x}_u^\top\cdot \textbf{x}_v) + \log(1-\sigma(\textbf{x}_u^\top\cdot \overline{\textbf{x}})),
        \label{eq:nie}
    \end{equation}
    where $\overline{\textbf{x}} = 1/(N+M)\sum_{u'\in{\textbf{U}\cup\textbf{I}}} \textbf{x}_{u'}$ denotes the average embedding of all the user-item embeddings. Different from node classification tasks, where we add more informative 1-hop neighbors from 2-hop neighbor set, due to the nature of the heterogeneous graph in recommendation tasks, the types of 1-hop neighbors are different from the types of the 2-hop neighbors. Thus, with this neighbor information evaluation function Eq.~(\ref{eq:nie}) defined for recommendation tasks, we select the most informative neighbors for each user~(item) node to construct the neighbor-enhanced heterogeneous user-item graph.
	\section{EXPERIMENT} 
	\label{sec:experiment}
	\subsection{Experimental Settings.}
	\noindent\textbf{Dataset Description.} We select four widely used datasets to demonstrate the power of the NEGCN, i.e., Cora, Citeseer, Pubmed, and Reddit~\cite{hamilton2017graphsage}. The size of the graphs scales from $O(10^3)$ to $O(10^5)$. We present the details of the four datasets in Table~\ref{tab:dataset}.
	\\
	\textbf{Edge classifier.} We train the edge classifier with the label-known nodes and edges from the training set. As shown in Table~\ref{tab:refine}, the number of positive edges is larger than that of negative edges. We perform negative sampling by randomly selecting different-label nodes. Especially, for feature prepossessing, we employ the parameter-free embedding $\textbf{A}^2 \textbf{X}$~\cite{wu2019sgc} as the input $\textbf{X}$ to train the edge classifier.
	For Cora, Citeseer and Pubmed, we refine the graph using both the adding process and the filtering process. For the adding process, we set the maximum number of 1-hop neighbors as $6$ for both Cora and Citeseer, and as $30$ for Pubmed. The statistical positive (negative) numbers of the original graph structure and the neighbor-enhanced graph structure are compared in Table \ref{tab:refine}.
	\\
	\textbf{Baseline models.} We select five representative GCN models as competitive baselines. The compared semi-supervised methods include the original GCN~\cite{kipf2016gcn}, GAT~\cite{velivckovic2017gat}, and SGC~\cite{wu2019sgc}. The compared supervised methods include GraphSAGE~\cite{hamilton2017graphsage} and ASGCN \cite{huang2018asgcn}. The semi-supervised methods only use a small part of nodes in the training set (presented as Semi-Supervised in Table~\ref{tab:dataset}) to optimize their parameters; while the supervised methods use all the nodes in the training set to optimize their parameters.
	For all the compared models, we fix the random seeds and use the early stopping strategy (using a window size of $30$ as suggested in~\cite{kipf2016gcn}) to generate the best performances. The presented performances are averaged over multiple runs to give a fair comparison.
	To reduce the performance fluctuation, we run these algorithms with 5 different random seeds and present the average result.
	
	\subsection{Experimental Results}    
	\begin{table}
		\renewcommand\arraystretch{1}
		\small
		\centering
		\caption{The statistics of the datasets.}
		\vskip -1em
		\setlength{\tabcolsep}{2.5mm}{
			\begin{tabular}{lcc}
				\toprule
				\textbf{Dataset} & \textbf{Nodes/Edges/Features} & \textbf{SemiTrain/Train/Val/Test} \\
				\midrule
				Cora  & 2,708/5,429/1,433 & 140/1,208/500/1,000 \\
				Citeseer & 3,327/4,732/3,703 & 120/1,812/500/1,000  \\
				Pubmed & 19,717/44,338/500 & 60/18,217/500/1,000  \\
				Reddit & 233.0K/11.6M/602 & -/152K/24K/55K\\
				\bottomrule
			\end{tabular}
		}
		\label{tab:dataset}
		\vskip -1em
	\end{table}
	
	\begin{table}
		\renewcommand\arraystretch{1}
		\small
		\centering
		\caption{The statistics of the origin graphs and the neighbor enhanced graphs.}
		\vskip -1em
		\setlength{\tabcolsep}{2.5mm}{
			\begin{tabular}{lcc}
				\toprule
				\textbf{Dataset} & \textbf{ori-Pos/ori-Neg/ori-Ratio} & \textbf{NE-Pos/NE-Neg/NE-Ratio} \\
				\midrule
				Cora  & 18.4K/3.2K/85\% & 20.3K/2.2K/90\% \\
				Citeseer & 6.8K/2.4K/74\% & 16.6K/3.7K/82\% \\
				Pubmed  & 71.1K/17.5K/80\% & 483.0K/20.4K/96\% \\
				Reddit & 18.1M/5.1M/78\% & 18.0M/1.0M/95\% \\
				\bottomrule
			\end{tabular}
		}
		\label{tab:refine}
		\vskip -1em
	\end{table}
	\begin{table}[t]
		\small
		\centering
		\caption{Accuracy of the NE-GCNs against the origin-GCNs}
		\vskip -1em
		\begin{tabular}{lccc}
			\toprule
			& \textbf{Cora} & \textbf{Citeseer} & \textbf{Pubmed} \\
			\midrule
			\textbf{Semi-Supervised Methods} &       &       &  \\\midrule
			origin-GCN & 0.8180$\pm$ 0.0065 & 0.7090$\pm$ 0.0032 & 0.7850$\pm$ 0.0049 \\
			NE-GCN & 0.8330$\pm$ 0.0057 & 0.7330$\pm$ 0.0045 & \textbf{0.8780$\pm$ 0.0045} \\
			\midrule
			origin-SGC & 0.8210$\pm$ 0.0005 & 0.7190$\pm$ 0.0012 & 0.7890 $\pm$ 0.0007\\
			NE-SGC & \textbf{0.8380$\pm$ 0.0005} & 0.7340 $\pm$ 0.0009& 0.8770 $\pm$ 0.0006\\
			\midrule
			origin-GAT & 0.8300$\pm$ 0.0075 & 0.7250 $\pm$ 0.0061& 0.7900 $\pm$ 0.0032\\
			NE-GAT & 0.8350 $\pm$ 0.0067 & \textbf{0.7360$\pm$ 0.0073} & 0.8690$\pm$ 0.0033 \\
			\midrule[0.8pt]
			\textbf{Supervised Methods} &       &       &  \\\midrule
			origin-SAGE & 0.8650 $\pm$ 0.0062 & 0.7850  $\pm$ 0.0081& 0.8830 $\pm$ 0.0114 \\
			NE-SAGE & 0.8840  $\pm$ 0.0048 & 0.8000 $\pm$ 0.0078 & 0.9070 $\pm$ 0.0098 \\
			\midrule
			origin-ASGCN & 0.8740  $\pm$ 0.0034 & 0.7960 $\pm$ 0.0018 & 0.9060 $\pm$ 0.0016 \\
			NE-ASGCN & \textbf{0.8880$\pm$ 0.0037 } & \textbf{0.8010$\pm$ 0.0022} & \textbf{0.9170$\pm$ 0.0015} \\
			\bottomrule
		\end{tabular}
		%\vskip -1em
		\label{tab:overall}
	\end{table}
	\begin{table}
		%\vskip -1em
		\small
		\centering
		\caption{Accuracy comparison on Reddit}
		\vskip -1em
		\begin{tabular}{ccccc}
			\toprule
			& \textbf{SGC} & \textbf{NE-SGC} & \textbf{ASGCN} & \textbf{NE-ASGCN} \\
			\midrule
			\textbf{Reddit} & 0.9488$\pm$ 0.0005 & 0.9540$\pm$ 0.0004 & 0.9627$\pm$ 0.0032 & \textbf{0.9758$\pm$ 0.0027 } \\
			\bottomrule
		\end{tabular}
		\label{tab:reddit}
		\vskip -1em
	\end{table}
	\begin{table}[t]
		\vskip -1.5em
		\small
		\centering
		\caption{Performance of the edge classifier.}
		\vskip -1em
		\begin{tabular}{lcccc}
			\toprule
			& \textbf{Cora} & \textbf{Citeseer} & \textbf{Pubmed} & \textbf{Reddit} \\
			\midrule
			$p$ & 85\%  & 89\%  & 92\%  & 98\% \\
			$q$ & 36\%  & 54\%  & 31\%  & 13\% \\
			$p-q$ & 49\%  & 34\%  & 61\%  & 85\% \\
			$p_{pre}$ & 93\%  & 82\%  & 93\%  & 97\% \\
			Accuracy & 86\%  & 75\%  & 87\%  & 96\% \\
			\bottomrule
		\end{tabular}
		\label{tab:edge}
		\vskip -1em
	\end{table} 
	
	Table~\ref{tab:refine} presents the statistic information before and after the neighbor enhanced process. The result shows that the graph refinement can improve the positive ratio remarkably.
	Table~\ref{tab:overall} compares GCN models and their NE enhanced versions on the Cora, Citeseer, and Pubmed. The results present that the proposed NEGCN can enhance classification performance considerably. Moreover, the results in Table~\ref{tab:refine} and Table~\ref{tab:overall} can verify Proposition~\ref{the:origin}.
	
	We now take a deeper look at the experimental results.
	For the semi-supervised methods, the result in  Table~\ref{tab:overall} shows that the original GAT outperforms the original GCN and the original SGC. Both the NE-SGC and the NE-GCN perform better than the original GAT, which indicates that the NEGCN framework is more effective than the attention mechanism. 
	Meanwhile, the NE-GAT also outperforms the original GAT, which indicates that NEGCN can complement the attention mechanism to reach a better performance. It is also noteworthy that the NE-GCN, the NE-SGC, and the NE-GAT outperform their original versions by almost $10$ percent on the Pubmed dataset.
	For the supervised methods, the NE-SAGE performs better than the original ASGCN. Moreover, as shown in Table~\ref{tab:reddit}, we also evaluate the performance of NEGCN on the Reddit dataset. The result shows that both NE-SGC and NE-ASGCN can achieve higher accuracy than the original SGC and the original ASGCN.
	\begin{figure*}[t]
		\begin{center}
			\centerline{\includegraphics[width=1.12\linewidth,trim=2cm 0 0 0,clip]{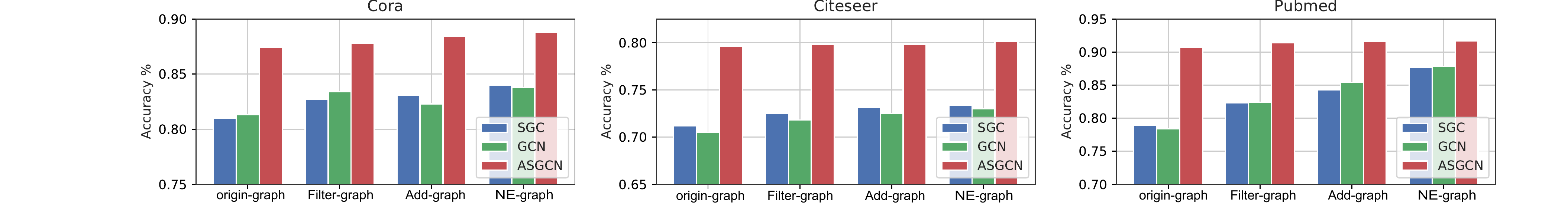}}
			\vskip -0.5em
			\caption{The performance of the GCN models under different graphs. Filter-graphs are generated by only using the filtering process; add-graphs are generated by only using the adding process; NE-graphs are generated by using both the adding and the filtering process.} 
			\label{fig:addfilt}
		\end{center}
		\vskip -2em
	\end{figure*}
	\begin{figure*}[htbp]
		% 		\vspace{-1em}
		\begin{center}
			\centerline{\includegraphics[width=1.03\linewidth,trim=0 0 0 0,clip]{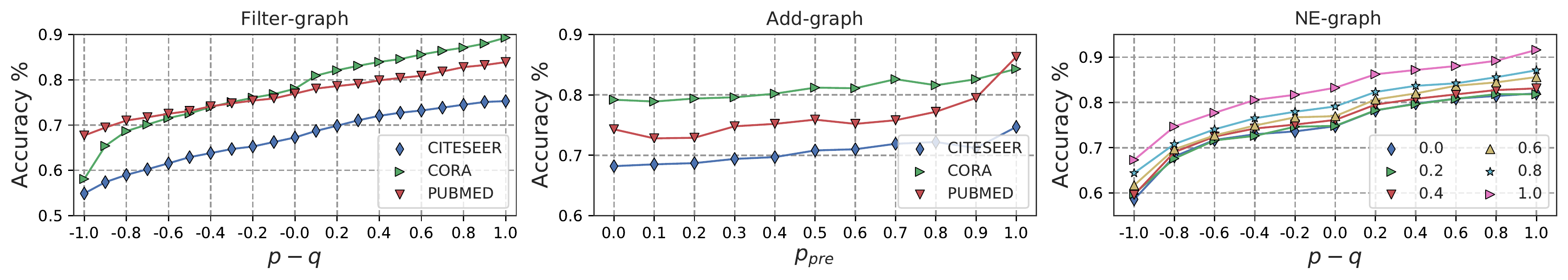}}
			
			\caption{The performance of NE-SGC when varying the values of $p-q$ and $p_{pre}$. The results on the left and the middle panel are generated with the filter-graph and the add-graph, respectively. The results on the right panel are generated with the NE-graph.}
			\label{fig:gr_predict}
		\end{center}
		\vskip -1em
	\end{figure*}
\vspace{0.2em}
	\subsection{Ablation Study}
	\textbf{Performance of the edge classifier.} 
	For each experimental dataset, we separate a validation set from the original training dataset. Then, we train the edge classifiers using the remained training set and evaluate the predicting accuracy of the edge classifier using the separated validation set. The result in Table \ref{tab:edge} shows that $p-q$ and $p_{pre}$ have a strong correlation with the evaluated predicting accuracy. Hence, one can select the best edge classifier according to their prediction accuracy evaluated on the validation set. 
	
	\smallskip\noindent\textbf{Influence from the graph enhancing process.} In Figure~\ref{fig:addfilt}, we evaluate the influence of the adding and filtering process by testing the neighbor enhanced GCN-models on the original graph, the filter-graph, the add-graph, and the NE-graph, respectively. In particular, the filter-graph only filters the negative edges; the add-graph only adds more positive edges; the NE-graph both filters the negative edges and adds more positive edges.
	The performance on the filter-graph and the add-graph show that both the filtering and the adding process can enhance the performance of the GCN models. The superior performance on the NE-graph shows that the filtering process and the adding process can complement each other to reach the best performance.
	
	\smallskip\noindent\textbf{Influence from the edge classifier.} In Figure \ref{fig:gr_predict}, we present the performance of NE-SGC when varying the values of $p$ , $q$ and $p_{pre}$. 
	Specifically, the sub-figure on the left panel shows that the performance of the NE-models is positively correlated to the value of $p-q$. When $p-q>0$, the accuracy of NE-SGC is higher than the original SGC; otherwise, the accuracy of NE-SGC is lower than the original SGC, which verifies Proposition~\ref{the:filter}.
	The sub-graph on the middle panel shows that the performance of NE-models is positively correlated to the value of $p_{pre}$. Specifically, the result shows that if $p_{pre} >R$, the accuracy of NE-SGC is higher than original SGC; otherwise, the accuracy of NE-SGC is lower, which verifies Proposition~\ref{the:add}.
	The sub-graph on the right panel shows how the performance of NE-SGC varies with the values of $p-q$ and $p_{pre}$. Each line corresponds to a specific value of $p_{pre}$. The result shows that the lines corresponding to different $p_{pre}$ values are almost in parallel, which indicates that the adding process and the filtering process have little influence on each other.
	
	\begin{table}[t]
		\vskip -1.5em
		\small
		\centering
		\caption{Performance under low positive ratio.}
		\vskip -1em
		\begin{tabular}{lccc}
			\toprule
			& \textbf{Cora} & \textbf{Citeseer} & \textbf{Pubmed} \\
			\midrule
			origin-GCN(low) & 0.3920 & 0.3820 & 0.5810 \\
			NE-GCN(low) & 0.6690 & 0.6650 & 0.8780 \\
			\midrule
			origin-GAT(low) & 0.4430 & 0.3580 & 0.6020 \\
			NE-GAT(low) & 0.6850 & 0.6750 & 0.8850 \\
			\bottomrule
		\end{tabular}
		\vskip -1em
		\label{tab:low-homophily-ratio}
	\end{table}
	
	\noindent\textbf{Performance under low positive ratio.}
	We add five different-label neighbors to each node in Cora, Citeseer and Pubmed graphs so as to artificially decrease their positive ratio from $85\%$, $74\%$, $80\%$ to $ 30\%$, $24\%$, $34\%$, respectively, and test the performance of GCN and GAT.
	We train the edge classifier~(using the raw features) which can improve the positive ratio back to $ 67\%$, $73\%$, $97\% $, respectively, and test the performance of NE-GCN and NE-GAT.
	The results in Table~\ref{tab:low-homophily-ratio} show that GCN and GAT generate a worse performance when the graph has a lower positive ratio. This may due to that the convolution and attention mechanism have limited effect when the nodes in a graph have few positive neighbors. 
	On the other hand, the superior performance of NE-GCN and NE-GAT proves that the NEGCN can help existing GCNs to achieve much better performance by refining the graph topology, especially when the graph has a low positive ratio.
	\subsection{Experiments on Recommendation Tasks}
	\noindent\textbf{Dataset Description.} 
	We adopt the publicly accessible recommendation dataset: Yelp to demonstrate the power of NEGCN on recommendation tasks. Yelp2018 is adopted from the 2018 edition of the Yelp challenge, where the items are specified as the local businesses like restaurants and bars. We randomly select 65\% of historical interactions of each user to learn the node embeddings with the off-the-shelf embedding methods, another 15\% of historical interactions to train the relevance function, and the remaining 20\% are used to validate/test the recommendation performance. In order to avoid the cold-starting problem, all users and items have at least one record in embedding learning~\cite{he2020lightgcn}.
	\begin{table}[ht]
    \small
    \centering
    \vspace{-0.5em}
    \caption{Comparison of sampling policies.}
    \vspace{-1em}
    \resizebox{0.8\columnwidth}{!}{
    	\setlength{\tabcolsep}{2mm}{
 \begin{tabular}{ccccc}
 \toprule
 Embeddings & Sampling & Yelp(PRE) & Yelp(REC) & Yelp(NDCG) \\
 \midrule
 \multirow{4}[4]{*}{Meta2Vec} & Random & 0.0376 & 0.0825 & 0.0705 \\
       & Intuitive & \underline{0.0415} & \underline{0.0908} & \underline{0.0781} \\
       & NEGCN & \textbf{0.043} & \textbf{0.0932} & \textbf{0.0807} \\
\cmidrule{2-5}       & Improvement & 3.61\% & 2.64\% & 3.33\% \\
 \midrule
 \multirow{4}[4]{*}{GRMF} & Random & 0.0325 & 0.071 & 0.059 \\
       & Intuitive & \underline{0.033} & \underline{0.0727} & \underline{0.0602} \\
       & NEGCN & \textbf{0.0342} & \textbf{0.0753} & \textbf{0.0623} \\
\cmidrule{2-5}       & Improvement & 3.64\% & 3.58\% & 3.49\% \\
 \midrule
 \multirow{4}[4]{*}{LightGCN} & Random & 0.0418 & 0.0892 & 0.0771 \\
       & Intuitive & \underline{0.0421} & \underline{0.0908} & \underline{0.0785} \\
       & NEGCN & \textbf{0.045} & \textbf{0.0954} & \textbf{0.0818} \\
\cmidrule{2-5}       & Improvement & 6.89\% & 5.07\% & 4.20\% \\
 \bottomrule
 \end{tabular}%

     \vspace{-1em}
    	}}
    	\label{tab:samp}
    \end{table}
	
	\noindent\textbf{Evaluation Metrics.}
	We adopt the popular all-ranking evaluation protocol, which has been widely used in recent literature~\cite{wang2019ngcf,he2020lightgcn}. For each user in the testing set, all the non-interacted items are treated as the negative items. Specially, we rank all the items in the dataset except the interacted items used in the training process, and then truncate the ranked list at 20 to calculate precision~(PRE@20), recall~(REC@20), and ndcg~(NDCG@20) metrics. We calculate all the three metrics for each testing user and reported the average PRE@20, REC@20, and NDCG@20 metric over all the testing users.
	
	\noindent\textbf{Embedding Methods.}
	We choose three representative embedding method to demonstrate the general power of NEGCN:
	\begin{itemize}
		\item MetaPath2Vec~\cite{dong2017metapath2vec} formalizes metapath-based random walks on heterogeneous graphs as corpus and then leverages skip-gram~\cite{mikolov2013word2vec} models to compute node embeddings. 
		\item GRMF~\cite{rao2015grmf} generates the traditional matrix factorization~\cite{koren2009MF} by adding the graph Laplacian regularizer to restrict connected nodes to have similar embeddings. We adjust the BPR loss~\cite{rendle2012bprmf} to enhance the recommendation performance for fair comparison. 
		\item LightGCN~\cite{he2020lightgcn} linearly propagates user/item information on the user-item interaction graph with a four-layer graph convolution, and uses the weighted sum of layer-wise embeddings as the final embeddings.
	\end{itemize}
	The embedding size is fixed to 64 for all models, and all the embedding methods are implemented with the official codes.
	
	\noindent\textbf{Performance under different sampling strategies}
	In~\autoref{tab:samp}, we compare three different neighbor sampling policies on the Yelp dataset:
	\textit{1.Random:}~Random walk-based sampling~\cite{ying2018graph}, which simulates random walks starting from each node and computes the L1-normalized visit count of neighbors visited by the random walk. \textit{2. Intuitive:}~First-order proximity-based sampling~\cite{fan2019meirec, Wang2019graph, wu2019dual, wang2019knowledge}, which examines the neighborhood similarity based on the edge weights~(e.g., number of clicks). \textit{3. NEGCN:}~Our proposed neighbor enhancing policies.
	Overall, the random sampling method generates the worst performance. Intuitive sampling outperforms random sampling since the edge weights can represent the importance of an edge. As showed by the improvement percentage, the NEGCN sampling method outperforms the best baselines by a significant margin, where the main reason is that NEGCN selects the most important neighbors for the recommendation task.

	\section{Conclusion}
    In this paper, we proposed the NEGCN framework, which at the first time emphasized the concept of neighbor quality and demonstrated that increasing the neighbor quality can enhance general GCN models for both node classification and recommendation tasks. Specifically, we introduced an efficient edge classifier to explicitly identify the useful neighbors and modify the graph structure to increase the neighbor quality. Extensive experiments verified that increasing the neighbor quality helped in enhancing a wide range of GCN models. 
    This paper shed insights on emphasizing the neighbor quality studies and thus provided a branch new research line of investigating how to boost the GCN models by refining the graph structure. 
    
    Future researches can be further conducted on investigating better edge classifiers. For example, researchers can develop more complicated edge classifiers by utilizing state-of-the-art link prediction models. In addition, GCN models are making significant progress in Nature Language Processing and Computer Vision domains. It is also suggested to study how to define the neighbor quality in these domains and increase the neighbor quality to enhance the GCN models.

% \bibliographystyle{plain}
%\bibliography{mybibfile}
\bibliography{ref}

\end{document}